\def\arxivbuild{}
\documentclass{article}
\usepackage{iclr2026_conference,times}
\usepackage[T1]{fontenc}
\usepackage{amsmath,amssymb,amsfonts,bm}
\usepackage{graphicx,booktabs,multirow,microtype}
\usepackage[dvipsnames]{xcolor}
\usepackage{algorithm,algpseudocode}
\usepackage[colorlinks=true,linkcolor=black,citecolor=teal!60!black,urlcolor=teal!60!black]{hyperref}
\usepackage{url}
\usepackage{tikz}
\usetikzlibrary{positioning,arrows.meta,calc,fit,shapes.misc,backgrounds}
\newcommand{\numStorms}{1,096}
\newcommand{\numFrames}{189,353}
\newcommand{\numWindows}{276}
\newcommand{\numValWindows}{166}
\newcommand{\numTestStorms}{46}
\newcommand{\paramTok}{16.6M}
\newcommand{\paramDyn}{129.2M}
\newcommand{\paramDirect}{2.21M}
\newcommand{\paramRes}{47.2M}

\newcommand{\paramShare}{56}
\newcommand{\paramRatio}{1.8}
\newcommand{\selTau}{1.2}
\newcommand{\selRho}{0.9}
\newcommand{\selGamma}{1}
\newcommand{\rawTau}{1.0}
\newcommand{\rawRho}{0.9}
\newcommand{\recTau}{1.2}
\newcommand{\recRho}{0.9}
\newcommand{\recGamma}{0.75}
\newcommand{\twMSE}{0.0242}
\newcommand{\twCRPS}{0.0839}
\newcommand{\twES}{1.475}
\newcommand{\twCold}{0.0115}

\newcommand{\twWC}{0.78}
\newcommand{\twMC}{0.25}

\newcommand{\recWC}{0.66}

\newcommand{\gaussCRPS}{0.0828}
\newcommand{\gaussES}{1.829}

\newcommand{\gaussWC}{0.78}
\newcommand{\gaussMC}{0.33}

\newcommand{\bootES}{1.474}

\newcommand{\bootWC}{0.21}

\newcommand{\rawMSE}{0.0291}

\newcommand{\dirMSE}{0.0242}
\newcommand{\dirMAE}{0.1164}

\newcommand{\wideMSE}{0.0242}

\newcommand{\seedSD}{0.00005}

\newcommand{\gainMSEraw}{17}
\newcommand{\gainMSEconv}{34}
\newcommand{\gainColdLo}{14}
\newcommand{\gainColdHi}{39}
\newcommand{\gainCfourLo}{21}
\newcommand{\gainCfourHi}{43}

\newcommand{\gainESgauss}{19}
\newcommand{\gapESrec}{0.6}
\newcommand{\gapCRPSgauss}{1}
\newcommand{\gainCRPSboot}{3}
\newcommand{\gainCRPSrec}{1}

\newcommand{\gapBootLo}{0.44}
\newcommand{\gapBootHi}{0.69}
\newcommand{\gapGaussLo}{-0.04}
\newcommand{\gapGaussHi}{0.04}
\newcommand{\gapRecLo}{0.07}
\newcommand{\gapRecHi}{0.17}
\newcommand{\imgGradWC}{0.75}
\newcommand{\partWC}{0.41}
\newcommand{\hrsColdAll}{44}

\newcommand{\wonColdMin}{63}
\newcommand{\wonCRPSrec}{87}

\newcommand{\seedColdLo}{14}
\newcommand{\seedColdHi}{39}
\newcommand{\nsColdLo}{10}
\newcommand{\nsColdHi}{35}
\newcommand{\nsWonColdMin}{59}
\newcommand{\gridWord}{seven}
\newcommand{\nsStateGain}{5}
\newcommand{\nsOtherMax}{1}
\newcommand{\diffCRPS}{0.0838}

\newcommand{\diffWC}{0.77}

\newcommand{\diffGamma}{1.5}
\newcommand{\paramDiff}{50.4M}

\newcommand{\gcES}{1.474}

\newcommand{\gcWC}{0.78}

\newcommand{\gcSigma}{4}
\newcommand{\gainColdDiff}{14}
\newcommand{\gainCfourDiff}{21}

\newcommand{\gapESdiff}{0.5}
\newcommand{\wonColdDiff}{63}
\newcommand{\wonCfourDiff}{74}
\newcommand{\phyMSE}{0.0246}
\newcommand{\tauMSE}{0.0259}

\newcommand{\gapDiffLo}{-0.03}
\newcommand{\gapDiffHi}{0.04}
\newcommand{\gapGCLo}{-0.04}
\newcommand{\gapGCHi}{0.04}

\newcommand{\model}{TyphoonWorld}
\newcommand{\x}{\bm{x}}
\newcommand{\z}{\bm{z}}
\newcommand{\cc}{\bm{c}}
\newcommand{\E}{\mathbb{E}}

\title{TyphoonWorld: A Residual Token World\\Model for Probabilistic Forecasting\\of Typhoon Imagery}
\ifdefined\arxivbuild
\author{Xiaoyu Chen\\Nanyang Business School\\Nanyang Technological University, Singapore\\xiaoyu.chen@ntu.edu.sg
\And Tianyi Li\\National Academy of Governance\\Beijing, China
\And Xuanyi Zhang\thanks{Corresponding author.}\\PBC School of Finance\\Tsinghua University, Beijing, China\\zhangxy.18@pbcsf.tsinghua.edu.cn}
\else
\author{Anonymous authors\\Paper under double-blind review}
\fi
\iclrfinalcopy

\begin{document}
\maketitle

\begin{abstract}
A satellite forecast of a typhoon should give a low-error picture of the next two days, together with an ensemble whose members are coherent fields and whose spread responds to the storm at hand. A small direct regression network already gives the lowest squared error on held-out storms, so a generative model has to earn its cost through the part of the future that the regression smooths away. We present \model{}, a discrete-token world model that generates only that part. Conditioned on tokens of the regression forecast, the latest frames and the best-track state at forecast time, it samples residual fields lead by lead, and members add these residuals to the regression center. We compare it under one protocol with per-pixel Gaussian networks, a diffusion model of the same residuals, residuals from other storms and an unconditional token world model, on \numWindows{} forecast windows from \numTestStorms{} held-out storms of Digital Typhoon. \model{} matches the squared error of the regression, lowers the CRPS of the area colder than 220\,K, a direct measure of deep convection, by \gainColdLo{} to \gainColdHi\% against every other ensemble, and ties its spread to forecast error as closely as the Gaussian network, with a correlation of \twWC{} across windows. It uses \paramShare\% of the parameters of the unconditional world model.

\end{abstract}

\section{Introduction}\label{sec:intro}
A forecaster reading the latest infrared image of a typhoon needs two things: a good estimate of how the cloud field will evolve over the next two days, and an ensemble that shows what else could happen to this particular storm. Geostationary imagery is the main observation over the open ocean, and Digital Typhoon \citep{kitamoto2023digitaltyphoon} now offers hourly images of more than a thousand western Pacific storms with best-track records, so learned models can address both needs.

Each available tool delivers part of this. A direct U-Net with \paramDirect{} parameters, trained for about a quarter of an hour, reaches an MSE of \dirMSE{} over 48 hours on held-out storms, below the published spatiotemporal predictors we test, yet it returns one smooth field per lead in which deep convective cloud is washed out. A per-pixel Gaussian output on the same network tracks the size of the error, and its samples are spatially independent noise. Adding forecast residuals from other storms gives coherent members whose spread ignores the storm at hand. A world model built from discrete tokens and factorized attention in the style of recent video world models \citep{bruce2024genie,franch2025gptcast} samples whole future fields that respond to the current state, but with \paramDyn{} dynamics parameters the mean of those samples reaches an MSE of only \rawMSE{}. Placing its anomalies around the regression forecast restores the squared error, and most of what the world model learned then goes unused, because the center already explains the predictable part of each future.

\begin{figure}[t]
\centering
\resizebox{\linewidth}{!}{%
\begin{tikzpicture}[font=\sffamily\footnotesize,
  panel/.style={draw=black!65,dashed,rounded corners=7pt,line width=0.5pt,inner sep=5pt},
  tile/.style={inner sep=0pt,draw=black!50,line width=0.3pt},
  blk/.style={draw=black!60,rounded corners=2pt,minimum height=5mm,align=center,line width=0.4pt},
  main/.style={-{Triangle[length=2.6mm,width=3mm]},line width=1.7pt},
  thin/.style={-{Stealth[length=1.6mm]},line width=0.5pt,black!70}]
\def\ts{1.55cm}
\node[tile] (o3) at (0.22,0.22) {\includegraphics[width=\ts]{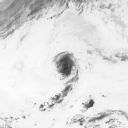}};
\node[tile] (o2) at (0.11,0.11) {\includegraphics[width=\ts]{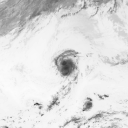}};
\node[tile] (o1) at (0,0) {\includegraphics[width=\ts]{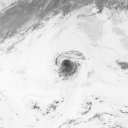}};
\node[below=1mm of o1,align=center] (ol) {last $K{=}8$ frames\\and state $\cc_{t_0}$};
\node[above=4.5mm of o1,font=\sffamily\footnotesize\bfseries] (ot) {Observed};
% center lane
\node[blk,fill=teal!18,minimum width=20mm] (un) at (3.5,1.25) {Direct U-Net};
\node[tile,right=7mm of un] (mu) {\includegraphics[width=\ts]{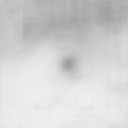}};
\node[below=0.5mm of mu] (mul) {center $\bm{\mu}$};
% residual lane
\node[blk,fill=orange!18,minimum width=17mm] (tk) at (3.15,-1.25) {Tokens of $\bm{\mu}_h$,\\$\x_{t_0}$, $\x_{t_0-6}$, $\cc_{t_0}$};
\node[blk,fill=orange!10,minimum width=24mm,right=4mm of tk,align=center] (tr) {Spatial $\cdot$ Temporal\\$\cdot$ Global attention};
\node[blk,fill=orange!18,minimum width=12mm,right=8mm of tr,align=center] (dec) {Residual\\decoder};
\node[tile] (r2) at (11.0,-1.14) {\includegraphics[width=\ts]{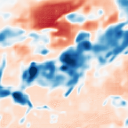}};
\node[tile] (r1) at (10.89,-1.25) {\includegraphics[width=\ts]{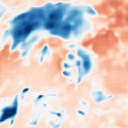}};
\node[below=0.5mm of r1,align=center] (rl) {residuals $\bm{\delta}^{(m)}$};
\draw[thin] (tk.east) -- (tr.west);
\draw[thin] (tr.east) -- node[above=0.6mm,font=\sffamily\scriptsize]{$\bm{w}^{(m)}_h$} (dec.west);
\draw[thin] (dec.east) -- (r1.west);
\draw[thin,rounded corners=3pt] (tr.south) -- ++(0,-0.38) -| ([xshift=3mm]tk.south);
\node[font=\sffamily\scriptsize,anchor=north] (fb) at ($(tk.south)!0.5!(tr.south)+(0,-0.42)$) {feed back residual codes};
\draw[thin,teal!60!black,rounded corners=3pt] (mul.south) -- ++(0,-0.35) -| ([xshift=4mm]tk.north);
% outputs
\node[blk,fill=blue!8,minimum width=15mm,minimum height=9mm,align=center] (rc) at (13.6,0) {Add zero-sum\\residuals\\Eq.~(\ref{eq:recenter})};
\node[tile] (m2) at (15.8,0.72) {\includegraphics[width=\ts]{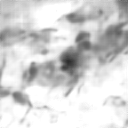}};
\node[tile] (m1) at (15.69,0.61) {\includegraphics[width=\ts]{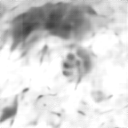}};
\node[above=0.5mm of m2,align=center] (ml) {members $\bm{r}^{(m)}$};
\node[tile] (sp) at (15.7,-1.25) {\includegraphics[width=\ts]{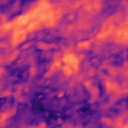}};
\node[below=0.5mm of sp] (sl) {spread};
\node[tile,right=2.5mm of sp] (er) {\includegraphics[width=\ts]{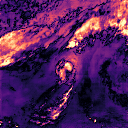}};
\node[below=0.5mm of er] (el) {actual error};
\draw[main,draw=teal!70!black] (o2.east)+(0.15,0.4) .. controls (1.5,1.25) .. (un.west);
\draw[main,draw=orange!70!black] (o2.east)+(0.15,-0.4) .. controls (1.5,-1.25) .. (tk.west);
\draw[thin] (un.east) -- (mu.west);
\draw[main,draw=teal!70!black] (mu.east) -- (rc.north west);
\draw[main,draw=orange!70!black] (r2.east)+(0.02,0.1) -- (rc.south west);
\draw[main,draw=black!70] (rc.east) -- (m1.west);
\draw[thin] (m1.south) -- (sp.north);
\draw[<->,>=Stealth,line width=0.5pt,black!70] ([yshift=3mm]sp.east) -- ([yshift=3mm]er.west);
\node[panel,fit=(un)(mu)(mul)] (pa) {};
\node[panel,fit=(tk)(tr)(dec)(r2)(r1)(rl)(fb),inner sep=5pt] (pb) {};
\node[panel,fit=(rc)(m2)(ml)(sp)(er)(sl)(el),inner sep=5pt] (pc) {};
\node[font=\sffamily\footnotesize\bfseries,anchor=south west] at (pa.north west) {(a) Regression center};
\node[font=\sffamily\footnotesize\bfseries,anchor=north west] at (pb.south west) {(b) Residual world model};
\node[font=\sffamily\footnotesize\bfseries,anchor=south west] at (pc.north west) {(c) Ensemble};
\end{tikzpicture}}
\caption{Overview of \model{}. A direct U-Net estimates the mean at every lead (a). A residual world model reads tokens of the center, two recent frames and the initial state, and samples residual codes lead by lead, which a residual decoder turns into fields (b). Members add zero-sum residuals to the center (c). All tiles come from one held-out window at a 24-hour lead; residual tiles show departures warmer than the center in red and colder in blue.}\label{fig:overview}
\end{figure}

\model{} gives the generative model only the part the regression leaves out. A direct U-Net supplies the center of the forecast. A residual tokenizer turns the difference between each future frame and the center into discrete codes, and a factorized transformer samples these codes lead by lead. At every lead it reads tokens of the center at that lead, tokens of the two most recent observed frames six hours apart and the best-track state at forecast time. Members add the sampled residuals to the center with zero mean across members, following the recentering practice of operational ensemble systems \citep{wang2013gsi}. Figure~\ref{fig:overview} shows the pipeline.

The evaluation covers every stage of each storm's life, from genesis to dissipation, on \numWindows{} forecast windows from \numTestStorms{} storms in seasons held out from training and validation. All centered ensembles share one construction, with amplitudes tuned on validation and twelve members scored as empirical distributions. \model{} matches the MSE of its center. Against every other ensemble it lowers the CRPS of the area colder than 220\,K by \gainColdLo{} to \gainColdHi\% and the CRPS of the area colder than 240\,K by \gainCfourLo{} to \gainCfourHi\%; the closest competitor is a diffusion model of the same residuals, which matches the pixel CRPS of \model{}. Its spread correlates with forecast error at \twWC{} across windows, level with the Gaussian network and far above residuals from other storms at \bootWC{}. The Gaussian network keeps a pixel CRPS \gapCRPSgauss\% lower, and the energy score of \model{} is \gapESrec\% above the best ensemble on that score. Recentered rollouts from the unconditional world model, with \paramRatio{} times the parameters of \model{}, fall behind it on deep convection and on the link between spread and error.

Our contributions are empirical and methodological. We show that a small direct regression network is more accurate on typhoon imagery than the published spatiotemporal predictors we test, which changes what a generative model must offer to justify its cost. We introduce a residual token world model that reads the regression forecast and samples only its error, which directs generative capacity to the convective structure that the regression smooths away. We also provide a like-for-like protocol that separates what each kind of ensemble offers.

\section{Related work}\label{sec:related}
Deterministic predictors dominate work on typhoon imagery. Digital Typhoon itself reports a ConvLSTM baseline and notes that predicted frames blur quickly \citep{kitamoto2023digitaltyphoon}, and later networks on the same data add physics-inspired modules, frequency branches or feature fusion \citep{yuan2025c2phynet,zhao2026phydcnet,li2026fusiontyphoonpredictor}. Related architectures for Himawari typhoon sequences use attention memories and multi-scale recurrence \citep{ren2024samnet,wang2023emsn}. General spatiotemporal predictors such as PredRNN, SimVP, TAU and Earthformer supply the standard designs \citep{wang2017predrnn,gao2022simvp,tan2023tau,gao2022earthformer}. Each of these systems emits a single trajectory, which leaves open the question of how far to trust it. We build on the same regression idea for the center and add a state-dependent ensemble around it.

Generative models address that question for precipitation and global weather. DGMR, NowcastNet and PreDiff sample radar futures \citep{ravuri2021dgmr,zhang2023nowcastnet,gao2023prediff}, and GenCast produces calibrated global ensembles \citep{price2025gencast}. Tropical-cyclone imagery has recently received diffusion and rectified-flow generators over pixels or continuous latents \citep{nath2023cascadeddiffusion,ren2025videodiffusiontc,zannat2026latentrectifiedflow}. Two-stage nowcasters are closest to our design. DiffCast and CasCast train a stochastic component on the residual of a deterministic forecaster \citep{yu2024diffcast,gong2024cascast}, and a recent infrared system pairs a U-Net mean with a diffusion residual \citep{afzaligorooh2026unetdiffusion}; these systems generate the residual with diffusion over pixels or continuous latents. \model{} keeps the same division of labor and generates the residual with a discrete-token autoregressive world model, which samples whole residual trajectories lead by lead and reads the regression forecast through its tokens at every lead.

Discrete token models give the generative half of our system. VQ-VAE and VQGAN established two-stage token modeling \citep{vandenoord2017vqvae,esser2021vqgan}, VideoGPT carried it to video \citep{yan2021videogpt}, and GAIA-1, iVideoGPT and Genie made rollouts action-conditioned or interactive \citep{hu2023gaia1,wu2024ivideogpt,bruce2024genie}. GPTCast and related work apply tokens and transformers to radar nowcasting, where sampling provides spread \citep{franch2025gptcast,meo2024nowcastinggpt}. Our dynamics model follows this line, with factorized attention as in Genie. On this task a small direct network has lower squared error than the sampled mean of a far larger token model, so we point the token model at the residual of the regression.

Verification practice motivates the evaluation. Squared error favors smooth forecasts through the double penalty \citep{mathieu2016beyondmse,subich2025doublepenalty}, and proper scores with spread diagnostics are the accepted way to judge ensembles \citep{gneiting2007scoring,ferro2014fair,fortin2014spread}. Recentering perturbed members on a better analysis is routine in operational ensemble systems \citep{wang2013gsi}, and our members follow the same construction around the regression center.

\section{Problem setup}\label{sec:setup}
\paragraph{Data.}
Digital Typhoon \citep{kitamoto2023digitaltyphoon} pairs hourly thermal-infrared images centered on each western North Pacific tropical cyclone with best-track records of central pressure, position, wind speed and intensity grade. The release we use covers 1978 to 2022. After removing sequences shorter than 12 frames, \numStorms{} storms and \numFrames{} frames remain. Each $512\times512$ image is block-averaged to $128\times128$ and mapped to $[0,1]$ with a fixed brightness-temperature window from 180\,K to 300\,K, so that a pixel value keeps the same physical meaning across storms and across time. Frames are centered on the best-track position at each hour, so all forecasts in this paper describe the cloud field in storm-relative coordinates, and the storm's track is outside the task. Splits follow storm year: storms up to 2018 train the models, 2019 and 2020 serve for validation and model selection, and 2021 and 2022 form the test set, which leaves a two-year gap between training and test storms. Table~\ref{tab:data} gives the counts.

\paragraph{Forecasting task.}
Let $\x_t\in[0,1]^{128\times128}$ denote the frame at hour $t$ and $\cc_t\in\mathbb{R}^{8}$ the best-track state, which holds normalized pressure, latitude, longitude, wind, grade and three clock features. At a forecast time $t_0$ a model receives the last $K=8$ frames $\x_{t_0-K+1:t_0}$ and must describe the next $H=48$ hourly frames. A deterministic model returns one trajectory. An ensemble model returns $M$ trajectories, whose mean is the point forecast and whose spread states the expected error at each lead and pixel. Every model either runs in closed loop or emits all leads at once, so no frame observed after $t_0$ enters a forecast.

\paragraph{Inputs at forecast time.}
Best-track records for hours after $t_0$ do not exist at forecast time, and post-season analysts use the same satellite imagery to produce them. Every regression baseline therefore sees frames alone. \model{} and the token world model of Section~\ref{sec:eval} also read the best-track state at $t_0$; during a rollout of the token world model its physical entries stay frozen at that value while clock features advance. An operational system would supply this state from a real-time analysis in place of the post-season record used here. No best-track value from after $t_0$ enters any forecast.

\section{The TyphoonWorld model}\label{sec:method}
\model{} has three parts, shown in Figure~\ref{fig:overview}. A direct regression network estimates the conditional mean of every future frame. A residual world model samples the difference between the future and that mean in a discrete token space. Members add zero-sum residual samples to the mean.

\subsection{Direct center}\label{sec:center}
A convolutional U-Net $f_\phi$ maps the $K$ observed frames to all $H$ leads at once and adds its output to the last observed frame,
\begin{equation}\label{eq:center}
\bm{\mu}_{t_0+h}=\x_{t_0}+f_\phi\bigl(\x_{t_0-K+1:t_0}\bigr)_h,\qquad
\mathcal{L}_{\mathrm{ctr}}=\E\Bigl[\tfrac{1}{H}\textstyle\sum_{h=1}^{H}\lVert\bm{\mu}_{t_0+h}-\x_{t_0+h}\rVert_2^2\Bigr].
\end{equation}
Each output channel targets the conditional mean at its lead. The network never consumes its own predictions, so errors do not compound along the forecast. The center has \paramDirect{} parameters and stays frozen once trained.

\subsection{Tokenizers}\label{sec:tokenizer}
A convolutional encoder $E$ maps a frame to a $16\times16$ grid of 256-dimensional vectors, and each vector is replaced by its nearest entry of a codebook $\{\bm{e}_k\}_{k=1}^{N}$ with $N=1024$,
\begin{equation}
z_{t}^{i}=\arg\min_{k}\,\lVert E(\x_t)^{i}-\bm{e}_k\rVert_2,\qquad i=1,\dots,256,
\end{equation}
where $\z_t\in\{1,\dots,N\}^{256}$ collects the codes of frame $t$. A mirrored decoder $D$ reconstructs $\tilde{\x}_t=D(\bm{e}_{\z_t})$. Training minimizes
\begin{equation}
\mathcal{L}_{\mathrm{tok}}=\lVert\x-\tilde{\x}\rVert_1+\lambda\bigl(1-\mathrm{SSIM}(\x,\tilde{\x})\bigr)+\beta\,\lVert E(\x)-\mathrm{sg}[\bm{e}_{\z}]\rVert_2^2,
\end{equation}
with $\lambda=0.3$, $\beta=0.25$ and stop-gradient $\mathrm{sg}[\cdot]$. Codes follow exponential moving averages of their assigned encoder outputs \citep{vandenoord2017vqvae,razavi2019generating}. Quasi-static clear-sky background otherwise captures most assignments early in training, so every 1,000 steps we replace codes whose usage falls below $10^{-3}$ with encoder outputs drawn from the current batch. This frame tokenizer has \paramTok{} parameters and describes observed frames and the center, whose codes at lead $h$ we write $\z^{\mu}_{h}$.

A residual tokenizer with the same architecture describes what the center leaves out. The residual of frame $t$ is $\bm{\delta}_t=\x_t-\bm{\mu}_t$, which we shift and scale to $\mathrm{clip}_{[0,1]}(0.5+\bm{\delta}_t/s)$ with $s=1.5$ so that it occupies the input range of the frame tokenizer. The residual tokenizer starts from the frame tokenizer's weights and is trained with the same loss on residuals of training storms at randomly drawn leads, with centers from the frozen regression network. Its codes at lead $h$ are $\bm{w}_h\in\{1,\dots,N\}^{256}$, and inverting the shift and scale after decoding returns a residual field.

\subsection{Residual dynamics}\label{sec:dynamics}
The residual world model $p_\theta$ predicts the residual codes of each lead from the residual codes of earlier leads and from conditions known at forecast time,
\begin{equation}\label{eq:res}
p_\theta\bigl(\bm{w}_{h}\mid\bm{w}_{<h},\bm{s}_{\le h}\bigr)=\prod_{i=1}^{256}p_\theta\bigl(w_{h}^{i}\mid\bm{w}_{<h},\bm{s}_{\le h}\bigr),\qquad
\bm{s}_{h}=\bigl(\z^{\mu}_{h},\z_{t_0},\z_{t_0-6},\cc_{t_0}\bigr),
\end{equation}
where $\z_{t_0}$ and $\z_{t_0-6}$ are codes of the last observed frame and of the frame six hours earlier. Lead $h$ enters as 256 tokens, one per grid position. Each token sums embeddings of the residual code of the previous lead, the center code for lead $h$ and the two observed-frame codes at that position, together with a projection of $\cc_{t_0}$, a lead embedding and a positional embedding; a start code takes the place of the previous residual at the first lead. Attention follows the spatiotemporal factorization of video world models \citep{bruce2024genie}: six spatial blocks attend within a lead, six temporal blocks attend causally across leads at a fixed grid position, and two global blocks let every token attend to all tokens of the current and earlier leads, over a sliding window of 16 leads. The model has width 512, eight heads and \paramRes{} parameters. Training minimizes label-smoothed cross-entropy on teacher-forced windows of residual codes from training storms,
\begin{equation}
\mathcal{L}_{\mathrm{res}}=-\E\Bigl[\textstyle\sum_{h}\sum_{i}\log p_\theta\bigl(w_{h}^{i}\mid\bm{w}_{<h},\bm{s}_{\le h}\bigr)\Bigr].
\end{equation}
Because each lead sees the center at that lead, the model spends its capacity on the distribution of the regression error, which changes with the storm, the lead and the predicted field.

\subsection{Ensemble construction}\label{sec:recenter}
A forecast samples Equation~\eqref{eq:res} lead by lead $M$ times, drawing each token from its distribution rescaled to temperature $\tau$ and truncated to nucleus mass $\rho$, written $q_{\tau,\rho}$ \citep{holtzman2020curious}, and decodes each trajectory with the residual tokenizer, which gives residual fields $\bm{\delta}^{(m)}_{t}$. Member $m$ is
\begin{equation}\label{eq:recenter}
\bm{r}^{(m)}_{t}=\mathrm{clip}_{[0,1]}\Bigl(\bm{\mu}_{t}+\gamma\bigl(\bm{\delta}^{(m)}_{t}-\bar{\bm{\delta}}_{t}\bigr)\Bigr),
\end{equation}
where $\bar{\bm{\delta}}_t$ is the member mean and $\gamma$ scales amplitude. Ensemble prediction systems recenter perturbed members on a better central estimate in the same way \citep{wang2013gsi}. The ensemble mean stays at $\bm{\mu}_t$ up to clipping, so the point forecast inherits the accuracy of the regression network, and each member carries a residual drawn for this storm given what the center predicts at that lead. Decoding parameters $(\tau,\rho)$ and $\gamma$ are the only quantities chosen after training, and validation windows alone determine them. Subtracting the member mean makes members dependent, which Section~\ref{sec:eval} accounts for in scoring. Algorithm~\ref{alg:rollout} states the procedure.

\begin{algorithm}[t]
\caption{Forecast with \model{}}\label{alg:rollout}
\begin{algorithmic}[1]
\Require frames $\x_{t_0-K+1:t_0}$, state $\cc_{t_0}$, members $M$, parameters $(\tau,\rho,\gamma)$
\State $\bm{\mu}_{t_0+1:t_0+H}\gets$ Equation~\eqref{eq:center}
\State $\z^{\mu}_{1:H},\z_{t_0},\z_{t_0-6}\gets$ frame-tokenizer codes of the center and of two observed frames
\For{$m=1,\dots,M$}
  \For{$h=1,\dots,H$}
    \State draw $\bm{w}^{(m)}_{h}\sim q_{\tau,\rho}\bigl(\cdot\mid\bm{w}^{(m)}_{<h},\bm{s}_{\le h}\bigr)$ over the last 16 leads
  \EndFor
  \State $\bm{\delta}^{(m)}_{t_0+1:t_0+H}\gets$ decode $\bm{w}^{(m)}_{1:H}$ and invert the shift and scale
\EndFor
\State $\bm{r}^{(m)}\gets$ Equation~\eqref{eq:recenter} for every $m$
\State \Return center $\bm{\mu}$ and members $\{\bm{r}^{(m)}\}$
\end{algorithmic}
\end{algorithm}

\section{Evaluation protocol}\label{sec:eval}
\paragraph{Forecast windows.}
Each test storm contributes one 56-hour window every 24 hours along its lifetime, from genesis to dissipation, which gives \numWindows{} windows across \numTestStorms{} storms. Validation storms give \numValWindows{} windows spaced 48 hours apart, on which all ensemble settings are chosen by lowest CRPS. For \model{} the search covers \gridWord{} decoding settings and six amplitudes and selects $\tau=\selTau$, $\rho=\selRho$ and $\gamma=\selGamma$, as Table~\ref{tab:grid} shows. All ensembles have $M=12$ members.

\paragraph{Deterministic baselines.}
Persistence repeats the last observed frame. ConvLSTM \citep{shi2015convlstm} is a three-layer recurrent model at full resolution, and a SimVP-style U-Net \citep{gao2022simvp} stacks the last eight frames; both predict one hour ahead, train with teacher forcing on mean squared error, and run in closed loop. PhyDNet \citep{leguen2020phydnet} represents the physics-guided recurrent designs used by recent predictors on Digital Typhoon \citep{yuan2025c2phynet,zhao2026phydcnet}: a recurrent cell whose kernels are constrained to act as partial derivatives runs beside a ConvLSTM branch at quarter resolution. TAU \citep{tan2023tau} maps eight frames to the next eight with temporal attention units and reaches 48 hours in chunks of eight, trained on all leads. We implement both from their papers. The direct U-Net of Section~\ref{sec:center}, trained at two widths and with three seeds at the smaller width, follows the one-shot design of SimVP. All regression baselines see frames only and train on the same storms for 12,000 steps with batch size 16.

\paragraph{Ensemble baselines.}
The Gaussian U-Net shares the architecture of the direct network, outputs a mean and a standard deviation at every pixel and lead, and trains on closed-form Gaussian CRPS, so it represents a strong per-pixel treatment of uncertainty at this scale. Its members add independent noise at each pixel. A correlated variant smooths the noise with a Gaussian kernel of width $\sigma$ pixels and carries it across leads with an autoregressive coefficient; pixel CRPS does not depend on this correlation, so the two are chosen by the validation energy score, which selects $\sigma=\gcSigma$, and the amplitude by CRPS. Diffusion residuals give the residual a continuous generative model in the manner of DiffCast and CasCast \citep{yu2024diffcast,gong2024cascast}. A conditional U-Net with \paramDiff{} parameters denoises all 48 residual leads at once from the center, the last eight frames and the best-track state at $t_0$, the inputs that \model{} reads, using a cosine schedule, $v$-prediction \citep{salimans2022progressive} and DDIM sampling \citep{song2021ddim}, with members formed by Equation~\eqref{eq:recenter} at $\gamma=\diffGamma$ selected on validation. The residual ensemble uses the same center $\bm{\mu}$ as \model{} and adds residual fields $\x-\bm{\mu}$ taken at the same lead from random validation windows; these fields have realistic amplitude, texture and spatial correlation by construction and ignore the current storm. The token world model measures what reading the center adds. It shares the frame tokenizer and models whole frames with the same factorized attention at width 768, with eight spatial, eight temporal and two global blocks and \paramDyn{} parameters, conditioned on the best-track state (Appendix~\ref{app:tokwm}). Token rollouts report its sampled ensemble around its own mean at the setting best for them on validation, $\tau=\rawTau$ and $\rho=\rawRho$. Recentered rollouts apply Equation~\eqref{eq:recenter} with $\bm{\delta}^{(m)}_t=\x^{(m)}_t-\bm{\mu}_t$ for its sampled frames $\x^{(m)}_t$, at $\tau=\recTau$, $\rho=\recRho$ and $\gamma=\recGamma$ selected on validation. The Gaussian, diffusion and residual ensembles follow the same construction: perturbations have zero mean across members and their amplitude is tuned on validation, so every centered ensemble shares one structure.

\paragraph{Metrics.}
MSE scores the point forecast at each lead. Members of a centered ensemble are dependent, so we score every ensemble as an empirical distribution \citep{gneiting2007scoring},
\begin{equation}
\mathrm{CRPS}=\frac{1}{M}\sum_{m}\bigl|\bm{r}^{(m)}-\x\bigr|-\frac{1}{2M^{2}}\sum_{m,m'}\bigl|\bm{r}^{(m)}-\bm{r}^{(m')}\bigr|,
\end{equation}
averaged over pixels. The score reduces to mean absolute error for a deterministic forecast. Storm-scale skill needs scores that see joint structure. The energy score applies the same form with Euclidean distances to $16\times16$ block means of each frame, and scalar CRPS applies to three storm summaries: the fractions of pixels colder than 220\,K and 240\,K, which track deep convection, and mean brightness in the central $32\times32$ box. To relate spread to error we use two correlations: at each lead across windows between mean spread and mean absolute error, and within each frame across pixels between the spread map and the absolute error map.

\paragraph{Statistical tests.}
Windows from one storm overlap, so inference treats the storm as the unit. Paired comparisons apply two-sided Wilcoxon signed-rank tests \citep{demsar2006statistical} to storm means across the \numTestStorms{} storms with Holm correction within each score, and the gap between two correlations uses a storm-level bootstrap with 2,000 resamples.

\section{Results}\label{sec:results}
\begin{table}[t]
\centering
\caption{Forecast skill on all test windows}\label{tab:main}
\small
\setlength{\tabcolsep}{3pt}
\begin{tabular}{lrrrrrrrrr}
\toprule
& & Point & \multicolumn{5}{c}{Proper scores$\downarrow$} & \multicolumn{2}{c}{Spread and error$\uparrow$} \\ \cmidrule(lr){3-3}\cmidrule(lr){4-8}\cmidrule(lr){9-10}
Model & Params & MSE$\downarrow$ & CRPS & Energy & 220\,K & 240\,K & Core & Window & Map \\
\midrule
Persistence & --- & 0.0459 & 0.1504 & 2.683 & 0.0194 & 0.0395 & 0.0823 & --- & --- \\
ConvLSTM & 0.52M & 0.0368 & 0.1329 & 2.567 & 0.0482 & 0.1369 & 0.2015 & --- & --- \\
SimVP-style, closed loop & 2.18M & 0.0442 & 0.1556 & 2.868 & 0.0493 & 0.1288 & 0.2226 & --- & --- \\
PhyDNet & 2.93M & 0.0246 & 0.1185 & 2.012 & 0.0413 & 0.0972 & 0.0698 & --- & --- \\
TAU, chunked & 41.3M & 0.0259 & 0.1195 & 2.090 & 0.0420 & 0.0987 & 0.0750 & --- & --- \\
Direct U-Net & 2.21M & \textbf{0.0242} & 0.1164 & 1.978 & 0.0405 & 0.0952 & 0.0669 & --- & --- \\
\midrule
Gaussian U-Net & 2.24M & 0.0245 & \textbf{0.0828} & 1.829 & 0.0173 & 0.0441 & 0.0650 & \textbf{0.78} & \textbf{0.33} \\
Gaussian U-Net, correlated & 2.24M & 0.0245 & \textbf{0.0828} & 1.474 & 0.0143 & 0.0375 & 0.0512 & \textbf{0.78} & \textbf{0.33} \\
Direct U-Net + residuals & 2.21M & 0.0243 & 0.0864 & 1.474 & 0.0172 & 0.0347 & \textbf{0.0496} & 0.21 & 0.12 \\
Diffusion residuals & 52.6M & \textbf{0.0242} & 0.0838 & 1.467 & 0.0134 & 0.0321 & 0.0499 & 0.77 & 0.27 \\
Token rollouts & 145.8M & 0.0291 & 0.0897 & 1.651 & 0.0159 & 0.0399 & 0.0689 & 0.69 & 0.31 \\
Recentered rollouts & 148.0M & \textbf{0.0242} & 0.0851 & \textbf{1.465} & 0.0187 & 0.0373 & 0.0502 & 0.66 & 0.19 \\
\model{} & 82.6M & \textbf{0.0242} & 0.0839 & 1.475 & \textbf{0.0115} & \textbf{0.0253} & 0.0501 & \textbf{0.78} & 0.25 \\
\bottomrule
\end{tabular}
\par\vspace{3pt}\parbox{\linewidth}{\footnotesize Notes: Scores average 276 windows from 46 storms over all 48 leads. Ensembles have twelve members and are scored as empirical distributions. Apart from token rollouts, every ensemble is zero-sum around its center with amplitude tuned on validation windows. Parameter counts include both tokenizers for \model{}, the tokenizer for token rollouts and the center for diffusion residuals. Deterministic rows use absolute error in place of CRPS and Euclidean error in place of the energy score. Energy is computed on $16\times16$ block means. The 220\,K and 240\,K columns are CRPS of the fraction of pixels colder than each threshold. Core is CRPS of mean brightness in the central $32\times32$ box. Window and Map are the correlations between spread and absolute error defined in Section~\ref{sec:eval}. Bold marks the best forecast model in each column.}
\end{table}

\subsection{Point forecast accuracy}\label{sec:res-point}
\begin{figure}[t]
\centering
\includegraphics[width=\linewidth]{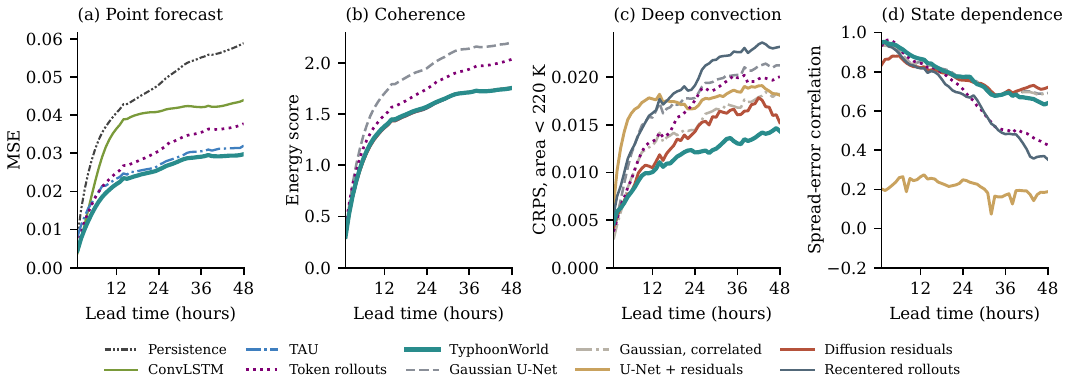}
\caption{Skill by lead time on all test windows. \model{} holds the error of the regression center in (a), stays level with the best energy scores in (b), has the lowest CRPS of the area colder than 220\,K at nearly every lead in (c), and keeps spread tied to forecast error in (d), where residuals from other storms average \bootWC{}.}\label{fig:lead}
\end{figure}

Table~\ref{tab:main} and Figure~\ref{fig:lead}a compare point forecasts. The direct U-Net reaches an MSE of \dirMSE{}, with a standard deviation of \seedSD{} across three training seeds, which is \gainMSEconv\% below ConvLSTM and \gainMSEraw\% below the mean of the token rollouts. PhyDNet comes closest at \phyMSE{}, and TAU reaches \tauMSE{}. ConvLSTM and the SimVP-style network train on one-step error and hold a small advantage in the first hours, then accumulate their own mistakes, whereas each output channel of the direct network targets the conditional mean at its lead. Widening the network to 8.77M parameters leaves MSE at \wideMSE{}. \model{} inherits this accuracy through its zero-sum construction, with an MSE of \twMSE{}. Token rollouts improve on persistence at every lead and stay above the direct network at every lead, as Table~\ref{tab:lead} details.

\subsection{Pixel-scale skill}\label{sec:res-prob}
All ensembles improve clearly on the deterministic center, whose absolute error is \dirMAE{}. The Gaussian U-Net optimizes pixel CRPS directly and reaches the best value at \gaussCRPS{} with independent or correlated noise. \model{} follows at \twCRPS{}, level with \diffCRPS{} for diffusion residuals, \gapCRPSgauss\% above the Gaussian network, \gainCRPSrec\% below recentered rollouts and \gainCRPSboot\% below the residual ensemble. It has the lower score on \wonCRPSrec\% of storms against recentered rollouts and on all storms against residuals, as Table~\ref{tab:tests} reports.

\subsection{Storm-scale skill}\label{sec:res-storm}
\begin{figure}[t]
\centering
\includegraphics[width=\linewidth]{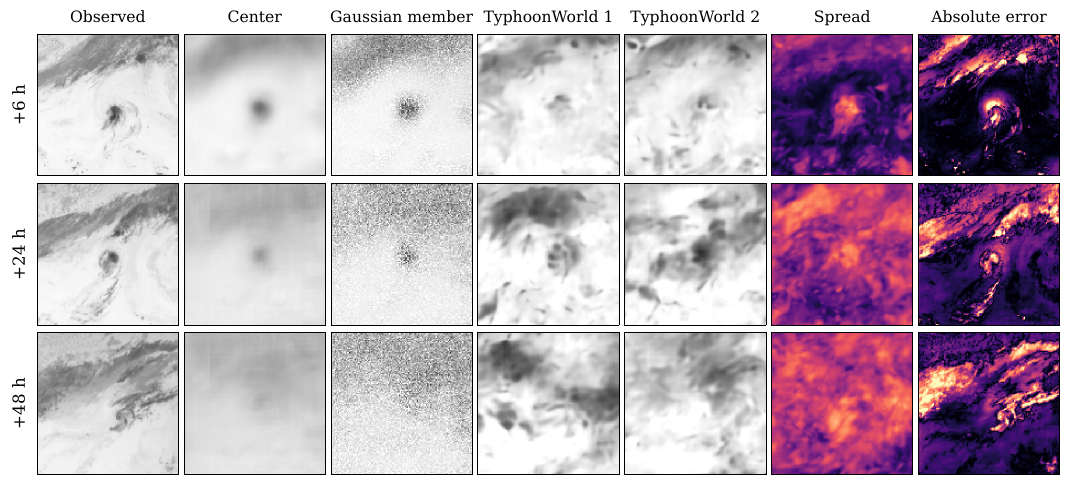}
\caption{Case study on a held-out storm. Rows are lead times. Spread and error refer to the twelve \model{} members and share one color scale. The window has the median CRPS difference between \model{} and the residual ensemble among windows after genesis.}\label{fig:case}
\end{figure}

Deep convection separates \model{} from every other ensemble. CRPS of the area colder than 220\,K is \twCold{}, which is \gainColdLo{} to \gainColdHi\% below the six alternatives; \model{} has the lower score on at least \wonColdMin\% of storms in each comparison and the lowest value at \hrsColdAll{} of 48 leads (Figure~\ref{fig:lead}c). At 240\,K the reduction ranges from \gainCfourLo{} to \gainCfourHi\%. Against diffusion residuals, the strongest alternative at 220\,K, the reductions are \gainColdDiff\% at 220\,K and \gainCfourDiff\% at 240\,K, with \model{} lower on \wonColdDiff\% and \wonCfourDiff\% of storms. These areas count the pixels that the regression blurs most, and the residual world model draws them for the storm at hand given the center it is correcting. Across three training seeds the reduction ranges from \seedColdLo{} to \seedColdHi\%. A variant trained without the best-track state, which conditions on frames alone as the regression baselines do, lowers the same score by \nsColdLo{} to \nsColdHi\% and wins on at least \nsWonColdMin\% of storms in each comparison; adding the state lowers the 220\,K score of \model{} by a further \nsStateGain\% and moves its MSE, pixel CRPS and energy score by at most \nsOtherMax\% (Table~\ref{tab:seeds}). On $16\times16$ block means the energy score of \model{} is \twES{}, level with \bootES{} for the residual ensemble and \gcES{} for the correlated Gaussian network, \gapESrec\% above recentered rollouts and \gapESdiff\% above diffusion residuals, and \gainESgauss\% below \gaussES{} for the Gaussian U-Net with independent noise, whose samples average out over each block (Figure~\ref{fig:lead}b). In mean brightness of the storm core, residuals and diffusion residuals score slightly below \model{}, and every centered ensemble with coherent members, the correlated Gaussian network included, stays well below the Gaussian network with independent noise. Figure~\ref{fig:case} shows a window chosen by rule as the median case: the center is a smooth field, a Gaussian member adds pixel noise, and \model{} members carry banded cold cloud and clear gaps that differ from member to member.

\subsection{State dependence of spread}\label{sec:res-flow}
A useful ensemble widens when the forecast is hard. Across windows at each lead, spread and mean absolute error correlate at \twWC{} for \model{}, against \recWC{} for recentered rollouts and \bootWC{} for the residual ensemble. Storm-level 95\% intervals for these two gaps run from \gapRecLo{} to \gapRecHi{} and from \gapBootLo{} to \gapBootHi{}, and \model{} leads both at every lead (Figure~\ref{fig:lead}d). The Gaussian U-Net reaches \gaussWC{}, its correlated variant \gcWC{} and diffusion residuals \diffWC{}; the intervals for their gaps to \model{}, from \gapGaussLo{} to \gapGaussHi{}, from \gapGCLo{} to \gapGCHi{} and from \gapDiffLo{} to \gapDiffHi{}, place all three level with it. Within a frame the Gaussian network has the higher pixelwise correlation between the spread map and the error map, at \gaussMC{} against \twMC{}, so a dedicated variance output remains the better tool for locating error pixel by pixel. Forecast difficulty is also visible in the input: the mean gradient of the last observed frame correlates with the error at \imgGradWC{}. After removing seven such image statistics from both quantities, the spread of \model{} keeps a partial correlation of \partWC{} with the error, so the residual world model contributes information beyond simple descriptors of the input. \model{} is thus the one ensemble that joins an accurate center, coherent members, the best deep-convection scores and spread as state dependent as the Gaussian network.

\section{Conclusion}\label{sec:conclusion}
\model{} lets a discrete-token world model generate only what a direct regression leaves unexplained. On held-out typhoons it matches the squared error of the regression network, lowers the CRPS of deep convective cloud area by \gainColdLo{} to \gainColdHi\% against every other ensemble tested, and ties its spread to forecast error as closely as a dedicated Gaussian variance network, with \paramShare\% of the parameters of an unconditional token world model whose recentered samples it improves on. The deep-convection gain holds across three training seeds and, at \nsColdLo{} to \nsColdHi\%, for a variant that conditions on frames alone. A per-pixel Gaussian network keeps a slightly lower pixel CRPS and locates error better within a frame, and a diffusion model of the same residuals matches the pixel and energy scores of \model{} within one percent, while \model{} lowers its 220\,K area CRPS by \gainColdDiff\%. Evaluation covers two seasons of one basin, and Digital Typhoon's other basins offer a direct extension. Because the regression network trains separately, a stronger center can take the place of the current one, with the residual tokenizer and world model retrained on its residuals.

\bibliographystyle{iclr2026_conference}
\bibliography{Refs}

\begin{thebibliography}{41}
\providecommand{\natexlab}[1]{#1}
\providecommand{\url}[1]{\texttt{#1}}
\expandafter\ifx\csname urlstyle\endcsname\relax
  \providecommand{\doi}[1]{doi: #1}\else
  \providecommand{\doi}{doi: \begingroup \urlstyle{rm}\Url}\fi

\bibitem[Afzali~Gorooh et~al.(2026)Afzali~Gorooh, Monache, Axisa, Sengupta,
  Zhang, and Ralph]{afzaligorooh2026unetdiffusion}
Vesta Afzali~Gorooh, Luca~Delle Monache, Duncan Axisa, Agniv Sengupta, Zhenhai
  Zhang, and Fred~Martin Ralph.
\newblock {Deterministic nowcasting of geostationary satellite infrared
  brightness temperature using 3D U-Net diffusion model}.
\newblock \emph{Scientific Reports}, 16\penalty0 (1):\penalty0 4191, 2026.
\newblock \doi{10.1038/s41598-025-34207-9}.

\bibitem[Bruce et~al.(2024)Bruce, Dennis, Edwards, Parker-Holder, Shi, Hughes,
  Lai, Mavalankar, Steigerwald, Apps, Aytar, Bechtle, Behbahani, Chan, Heess,
  Gonzalez, Osindero, Ozair, Reed, Zhang, Zolna, Clune, Freitas, Singh, and
  Rocktäschel]{bruce2024genie}
Jake Bruce, Michael~D Dennis, Ashley Edwards, Jack Parker-Holder, Yuge Shi,
  Edward Hughes, Matthew Lai, Aditi Mavalankar, Richie Steigerwald, Chris Apps,
  Yusuf Aytar, Sarah Maria~Elisabeth Bechtle, Feryal Behbahani, Stephanie~C.Y.
  Chan, Nicolas Heess, Lucy Gonzalez, Simon Osindero, Sherjil Ozair, Scott
  Reed, Jingwei Zhang, Konrad Zolna, Jeff Clune, Nando~De Freitas, Satinder
  Singh, and Tim Rocktäschel.
\newblock {Genie: Generative Interactive Environments}.
\newblock In \emph{International Conference on Machine Learning}, volume 235 of
  \emph{Proceedings of Machine Learning Research}, pp.\  4603--4623, 2024.
\newblock URL \url{https://proceedings.mlr.press/v235/bruce24a.html}.

\bibitem[Demšar(2006)]{demsar2006statistical}
Janez Demšar.
\newblock {Statistical Comparisons of Classifiers over Multiple Data Sets}.
\newblock \emph{Journal of Machine Learning Research}, 7\penalty0 (1):\penalty0
  1--30, 2006.
\newblock URL \url{https://jmlr.org/papers/v7/demsar06a.html}.

\bibitem[Esser et~al.(2021)Esser, Rombach, and Ommer]{esser2021vqgan}
Patrick Esser, Robin Rombach, and Bjorn Ommer.
\newblock {Taming Transformers for High-Resolution Image Synthesis}.
\newblock In \emph{2021 IEEE/CVF Conference on Computer Vision and Pattern
  Recognition (CVPR)}, pp.\  12868--12878, 2021.
\newblock \doi{10.1109/cvpr46437.2021.01268}.

\bibitem[Ferro(2014)]{ferro2014fair}
C.~A.~T. Ferro.
\newblock {Fair scores for ensemble forecasts}.
\newblock \emph{Quarterly Journal of the Royal Meteorological Society},
  140\penalty0 (683):\penalty0 1917--1923, 2014.
\newblock \doi{10.1002/qj.2270}.

\bibitem[Fortin et~al.(2014)Fortin, Abaza, Anctil, and
  Turcotte]{fortin2014spread}
V.~Fortin, M.~Abaza, F.~Anctil, and R.~Turcotte.
\newblock {Why Should Ensemble Spread Match the RMSE of the Ensemble Mean?}
\newblock \emph{Journal of Hydrometeorology}, 15\penalty0 (4):\penalty0
  1708--1713, 2014.
\newblock \doi{10.1175/jhm-d-14-0008.1}.

\bibitem[Franch et~al.(2025)Franch, Tomasi, Wanjari, Poli, Cardinali, Alberoni,
  and Cristoforetti]{franch2025gptcast}
Gabriele Franch, Elena Tomasi, Rishabh Wanjari, Virginia Poli, Chiara
  Cardinali, Pier~Paolo Alberoni, and Marco Cristoforetti.
\newblock {GPTCast: a weather language model for precipitation nowcasting}.
\newblock \emph{Geoscientific Model Development}, 18\penalty0 (16):\penalty0
  5351--5371, 2025.
\newblock \doi{10.5194/gmd-18-5351-2025}.

\bibitem[Gao et~al.(2022{\natexlab{a}})Gao, Tan, Wu, and Li]{gao2022simvp}
Zhangyang Gao, Cheng Tan, Lirong Wu, and Stan~Z. Li.
\newblock {SimVP: Simpler yet Better Video Prediction}.
\newblock In \emph{2022 IEEE/CVF Conference on Computer Vision and Pattern
  Recognition (CVPR)}, pp.\  3160--3170, 2022{\natexlab{a}}.
\newblock \doi{10.1109/cvpr52688.2022.00317}.

\bibitem[Gao et~al.(2022{\natexlab{b}})Gao, Shi, Wang, Zhu, Wang, Li, and
  Yeung]{gao2022earthformer}
Zhihan Gao, Xingjian Shi, Hao Wang, Yi~Zhu, Yuyang~(Bernie) Wang, Mu~Li, and
  Dit-Yan Yeung.
\newblock {Earthformer: Exploring Space-Time Transformers for Earth System
  Forecasting}.
\newblock In \emph{Advances in Neural Information Processing Systems},
  volume~35, pp.\  25390--25403, 2022{\natexlab{b}}.
\newblock \doi{10.52202/068431-1841}.
\newblock URL
  \url{https://proceedings.neurips.cc/paper_files/paper/2022/hash/a2affd71d15e8fedffe18d0219f4837a-Abstract-Conference.html}.

\bibitem[Gao et~al.(2023)Gao, Shi, Han, Wang, Jin, Maddix, Zhu, Li, and
  Wang]{gao2023prediff}
Zhihan Gao, Xingjian Shi, Boran Han, Hao Wang, Xiaoyong Jin, Danielle Maddix,
  Yi~Zhu, Mu~Li, and Yuyang~(Bernie) Wang.
\newblock {PreDiff: Precipitation Nowcasting with Latent Diffusion Models}.
\newblock In \emph{Advances in Neural Information Processing Systems},
  volume~36, pp.\  78621--78656, 2023.
\newblock \doi{10.52202/075280-3439}.
\newblock URL
  \url{https://proceedings.neurips.cc/paper_files/paper/2023/hash/f82ba6a6b981fbbecf5f2ee5de7db39c-Abstract-Conference.html}.

\bibitem[Gneiting \& Raftery(2007)Gneiting and Raftery]{gneiting2007scoring}
Tilmann Gneiting and Adrian~E Raftery.
\newblock {Strictly Proper Scoring Rules, Prediction, and Estimation}.
\newblock \emph{Journal of the American Statistical Association}, 102\penalty0
  (477):\penalty0 359--378, 2007.
\newblock \doi{10.1198/016214506000001437}.

\bibitem[Gong et~al.(2024)Gong, Bai, Ye, Xu, Liu, Dai, Yang, and
  Ouyang]{gong2024cascast}
Junchao Gong, Lei Bai, Peng Ye, Wanghan Xu, Na~Liu, Jianhua Dai, Xiaokang Yang,
  and Wanli Ouyang.
\newblock {CasCast: Skillful High-resolution Precipitation Nowcasting via
  Cascaded Modelling}.
\newblock In \emph{International Conference on Machine Learning}, volume 235 of
  \emph{Proceedings of Machine Learning Research}, pp.\  15809--15822, 2024.
\newblock URL \url{https://proceedings.mlr.press/v235/gong24a.html}.

\bibitem[Holtzman et~al.(2020)Holtzman, Buys, Du, Forbes, and
  Choi]{holtzman2020curious}
Ari Holtzman, Jan Buys, Li~Du, Maxwell Forbes, and Yejin Choi.
\newblock The curious case of neural text degeneration.
\newblock In \emph{International Conference on Learning Representations}, 2020.

\bibitem[Hu et~al.(2023)Hu, Russell, Yeo, Murez, Fedoseev, Kendall, Shotton,
  and Corrado]{hu2023gaia1}
Anthony Hu, Lloyd Russell, Hudson Yeo, Zak Murez, George Fedoseev, Alex
  Kendall, Jamie Shotton, and Gianluca Corrado.
\newblock {GAIA-1: A Generative World Model for Autonomous Driving}, 2023.

\bibitem[Kitamoto et~al.(2023)Kitamoto, Hwang, Vuillod, Gautier, Tian, and
  Clanuwat]{kitamoto2023digitaltyphoon}
Asanobu Kitamoto, Jared Hwang, Bastien Vuillod, Lucas Gautier, Yingtao Tian,
  and Tarin Clanuwat.
\newblock {Digital Typhoon: Long-term Satellite Image Dataset for the
  Spatio-Temporal Modeling of Tropical Cyclones}.
\newblock In \emph{Advances in Neural Information Processing Systems},
  volume~36, pp.\  40623--40636, 2023.
\newblock \doi{10.52202/075280-1768}.
\newblock URL
  \url{https://proceedings.neurips.cc/paper_files/paper/2023/hash/7fc36bce5de315751001981baaf4751a-Abstract-Datasets_and_Benchmarks.html}.

\bibitem[{Le Guen} \& Thome(2020){Le Guen} and Thome]{leguen2020phydnet}
Vincent {Le Guen} and Nicolas Thome.
\newblock {Disentangling Physical Dynamics From Unknown Factors for
  Unsupervised Video Prediction}.
\newblock In \emph{2020 IEEE/CVF Conference on Computer Vision and Pattern
  Recognition (CVPR)}, pp.\  11471--11481, 2020.
\newblock \doi{10.1109/cvpr42600.2020.01149}.

\bibitem[Li et~al.(2026)Li, Liu, Liu, and Liu]{li2026fusiontyphoonpredictor}
Haipeng Li, Jun Liu, Yan Liu, and Zelin Liu.
\newblock {FusionTyphoonPredictor: Dual-Branch Enhanced Spatiotemporal
  Prediction for Typhoon Cloud Images}.
\newblock \emph{Atmosphere}, 17\penalty0 (6):\penalty0 536, 2026.
\newblock \doi{10.3390/atmos17060536}.

\bibitem[Mathieu et~al.(2016)Mathieu, Couprie, and LeCun]{mathieu2016beyondmse}
Michaël Mathieu, Camille Couprie, and Yann LeCun.
\newblock {Deep multi-scale video prediction beyond mean square error}.
\newblock In \emph{International Conference on Learning Representations
  (ICLR)}, 2016.
\newblock URL \url{https://openreview.net/forum?id=ozp6HVYf7c}.

\bibitem[Meo et~al.(2024)Meo, Roy, Lică, Yin, Che, Wang, Imhoff, Uijlenhoet,
  and Dauwels]{meo2024nowcastinggpt}
Cristian Meo, Ankush Roy, Mircea Lică, Junzhe Yin, Zeineb~Bou Che, Yanbo Wang,
  Ruben Imhoff, Remko Uijlenhoet, and Justin Dauwels.
\newblock {Extreme Precipitation Nowcasting using Transformer-based Generative
  Models}, 2024.

\bibitem[Nath et~al.(2023)Nath, Shukla, Wang, and
  Quilodrán-Casas]{nath2023cascadeddiffusion}
Pritthijit Nath, Pancham Shukla, Shuai Wang, and César Quilodrán-Casas.
\newblock {Forecasting Tropical Cyclones with Cascaded Diffusion Models}, 2023.
\newblock Accepted for poster presentation at the ICLR 2024 workshop on
  Tackling Climate Change with Machine Learning.

\bibitem[Price et~al.(2025)Price, Sanchez-Gonzalez, Alet, Andersson, El-Kadi,
  Masters, Ewalds, Stott, Mohamed, Battaglia, Lam, and
  Willson]{price2025gencast}
Ilan Price, Alvaro Sanchez-Gonzalez, Ferran Alet, Tom~R. Andersson, Andrew
  El-Kadi, Dominic Masters, Timo Ewalds, Jacklynn Stott, Shakir Mohamed, Peter
  Battaglia, Remi Lam, and Matthew Willson.
\newblock {Probabilistic weather forecasting with machine learning}.
\newblock \emph{Nature}, 637\penalty0 (8044):\penalty0 84--90, 2025.
\newblock \doi{10.1038/s41586-024-08252-9}.

\bibitem[Ravuri et~al.(2021)Ravuri, Lenc, Willson, Kangin, Lam, Mirowski,
  Fitzsimons, Athanassiadou, Kashem, Madge, Prudden, Mandhane, Clark, Brock,
  Simonyan, Hadsell, Robinson, Clancy, Arribas, and Mohamed]{ravuri2021dgmr}
Suman Ravuri, Karel Lenc, Matthew Willson, Dmitry Kangin, Remi Lam, Piotr
  Mirowski, Megan Fitzsimons, Maria Athanassiadou, Sheleem Kashem, Sam Madge,
  Rachel Prudden, Amol Mandhane, Aidan Clark, Andrew Brock, Karen Simonyan,
  Raia Hadsell, Niall Robinson, Ellen Clancy, Alberto Arribas, and Shakir
  Mohamed.
\newblock {Skilful precipitation nowcasting using deep generative models of
  radar}.
\newblock \emph{Nature}, 597\penalty0 (7878):\penalty0 672--677, 2021.
\newblock \doi{10.1038/s41586-021-03854-z}.

\bibitem[Razavi et~al.(2019)Razavi, van~den Oord, and
  Vinyals]{razavi2019generating}
Ali Razavi, Aaron van~den Oord, and Oriol Vinyals.
\newblock Generating diverse high-fidelity images with {VQ-VAE-2}.
\newblock In \emph{Advances in Neural Information Processing Systems}, 2019.

\bibitem[Ren et~al.(2024)Ren, Ye, Wang, Xiao, and Liu]{ren2024samnet}
Yanzhao Ren, Jinyuan Ye, Xiaochuan Wang, Fengjin Xiao, and Ruijun Liu.
\newblock {SAM-Net: Spatio-Temporal Sequence Typhoon Cloud Image Prediction Net
  with Self-Attention Memory}.
\newblock \emph{Remote Sensing}, 16\penalty0 (22):\penalty0 4213, 2024.
\newblock \doi{10.3390/rs16224213}.

\bibitem[Ren et~al.(2025)Ren, Nath, and Shukla]{ren2025videodiffusiontc}
Zhibo Ren, Pritthijit Nath, and Pancham Shukla.
\newblock {Improving Tropical Cyclone Forecasting With Video Diffusion Models},
  2025.
\newblock Recommended for spotlight presentation at the ICLR 2025 workshop on
  Tackling Climate Change with Machine Learning.

\bibitem[Salimans \& Ho(2022)Salimans and Ho]{salimans2022progressive}
Tim Salimans and Jonathan Ho.
\newblock {Progressive Distillation for Fast Sampling of Diffusion Models}.
\newblock In \emph{International Conference on Learning Representations}, 2022.

\bibitem[Shi et~al.(2015)Shi, Chen, Wang, Yeung, kin Wong, and chun
  Woo]{shi2015convlstm}
Xingjian Shi, Zhourong Chen, Hao Wang, Dit-Yan Yeung, Wai kin Wong, and Wang
  chun Woo.
\newblock {Convolutional LSTM Network: A Machine Learning Approach for
  Precipitation Nowcasting}.
\newblock In \emph{Advances in Neural Information Processing Systems},
  volume~28, 2015.
\newblock URL
  \url{https://proceedings.neurips.cc/paper_files/paper/2015/hash/07563a3fe3bbe7e3ba84431ad9d055af-Abstract.html}.

\bibitem[Song et~al.(2021)Song, Meng, and Ermon]{song2021ddim}
Jiaming Song, Chenlin Meng, and Stefano Ermon.
\newblock {Denoising Diffusion Implicit Models}.
\newblock In \emph{International Conference on Learning Representations}, 2021.

\bibitem[Subich et~al.(2025)Subich, Husain, Separovic, and
  Yang]{subich2025doublepenalty}
Christopher Subich, Syed~Zahid Husain, Leo Separovic, and Jing Yang.
\newblock {Fixing the Double Penalty in Data-Driven Weather Forecasting Through
  a Modified Spherical Harmonic Loss Function}.
\newblock In \emph{International Conference on Machine Learning}, volume 267 of
  \emph{Proceedings of Machine Learning Research}, pp.\  57191--57211, 2025.
\newblock URL \url{https://proceedings.mlr.press/v267/subich25a.html}.

\bibitem[Tan et~al.(2023)Tan, Gao, Wu, Xu, Xia, Li, and Li]{tan2023tau}
Cheng Tan, Zhangyang Gao, Lirong Wu, Yongjie Xu, Jun Xia, Siyuan Li, and
  Stan~Z. Li.
\newblock {Temporal Attention Unit: Towards Efficient Spatiotemporal Predictive
  Learning}.
\newblock In \emph{2023 IEEE/CVF Conference on Computer Vision and Pattern
  Recognition (CVPR)}, pp.\  18770--18782, 2023.
\newblock \doi{10.1109/cvpr52729.2023.01800}.

\bibitem[van~den Oord et~al.(2017)van~den Oord, Vinyals, and koray
  kavukcuoglu]{vandenoord2017vqvae}
Aaron van~den Oord, Oriol Vinyals, and koray kavukcuoglu.
\newblock {Neural Discrete Representation Learning}.
\newblock In \emph{Advances in Neural Information Processing Systems},
  volume~30, 2017.
\newblock URL
  \url{https://proceedings.neurips.cc/paper_files/paper/2017/hash/7a98af17e63a0ac09ce2e96d03992fbc-Abstract.html}.

\bibitem[Wang et~al.(2023)Wang, Qin, Zhang, Wang, Du, and Wang]{wang2023emsn}
Xin Wang, Mengjiao Qin, Zhe Zhang, Yuanyuan Wang, Zhenhong Du, and Nan Wang.
\newblock {Typhoon cloud image prediction based on enhanced multi-scale deep
  neural network}.
\newblock \emph{Frontiers in Marine Science}, 9:\penalty0 956813, 2023.
\newblock \doi{10.3389/fmars.2022.956813}.

\bibitem[Wang et~al.(2013)Wang, Parrish, Kleist, and Whitaker]{wang2013gsi}
Xuguang Wang, David Parrish, Daryl Kleist, and Jeffrey Whitaker.
\newblock {GSI} {3DVar}-based ensemble-variational hybrid data assimilation for
  {NCEP} global forecast system: Single-resolution experiments.
\newblock \emph{Monthly Weather Review}, 141\penalty0 (11):\penalty0
  4098--4117, 2013.
\newblock \doi{10.1175/MWR-D-12-00141.1}.

\bibitem[Wang et~al.(2017)Wang, Long, Wang, Gao, and Yu]{wang2017predrnn}
Yunbo Wang, Mingsheng Long, Jianmin Wang, Zhifeng Gao, and Philip~S. Yu.
\newblock {PredRNN: Recurrent Neural Networks for Predictive Learning using
  Spatiotemporal LSTMs}.
\newblock In \emph{Advances in Neural Information Processing Systems},
  volume~30, 2017.
\newblock URL
  \url{https://proceedings.neurips.cc/paper_files/paper/2017/hash/e5f6ad6ce374177eef023bf5d0c018b6-Abstract.html}.

\bibitem[Wu et~al.(2024)Wu, Yin, Feng, He, Li, Hao, and Long]{wu2024ivideogpt}
Jialong Wu, Shaofeng Yin, Ningya Feng, Xu~He, Dong Li, Jianye Hao, and
  Mingsheng Long.
\newblock {iVideoGPT: Interactive VideoGPTs are Scalable World Models}.
\newblock In \emph{Advances in Neural Information Processing Systems},
  volume~37, pp.\  68082--68119, 2024.
\newblock \doi{10.52202/079017-2173}.
\newblock URL
  \url{https://proceedings.neurips.cc/paper_files/paper/2024/hash/7dbb5bfab324e3b86af9bd0df15498dd-Abstract-Conference.html}.

\bibitem[Yan et~al.(2021)Yan, Zhang, Abbeel, and Srinivas]{yan2021videogpt}
Wilson Yan, Yunzhi Zhang, Pieter Abbeel, and Aravind Srinivas.
\newblock {VideoGPT: Video Generation using VQ-VAE and Transformers}, 2021.

\bibitem[Yu et~al.(2024)Yu, Li, Ye, Zhang, Luo, Dai, Wang, and
  Chen]{yu2024diffcast}
Demin Yu, Xutao Li, Yunming Ye, Baoquan Zhang, Chuyao Luo, Kuai Dai, Rui Wang,
  and Xunlai Chen.
\newblock {DiffCast: A Unified Framework via Residual Diffusion for
  Precipitation Nowcasting}.
\newblock In \emph{2024 IEEE/CVF Conference on Computer Vision and Pattern
  Recognition (CVPR)}, pp.\  27758--27767, 2024.
\newblock \doi{10.1109/cvpr52733.2024.02622}.

\bibitem[Yuan et~al.(2025)Yuan, Zhao, Yu, Lu, Xia, Liu, Wang, and
  Wang]{yuan2025c2phynet}
Jiawei Yuan, Liling Zhao, Runling Yu, Xiaoqin Lu, Min Xia, Yi~Liu, Yuru Wang,
  and Xinyue Wang.
\newblock {A Physics-Enhanced Network for Predicting Sequential Satellite
  Images of Typhoon Clouds}.
\newblock \emph{IEEE Journal of Selected Topics in Applied Earth Observations
  and Remote Sensing}, 18:\penalty0 16798--16815, 2025.
\newblock \doi{10.1109/jstars.2025.3574201}.

\bibitem[Zannat \& Ahsan(2026)Zannat and Ahsan]{zannat2026latentrectifiedflow}
Meheru Zannat and Sk. Md.~Masudul Ahsan.
\newblock {Tropical Cyclone Forecasting via Latent Rectified Flow using
  Satellite Imagery and Atmospheric Fields}, 2026.

\bibitem[Zhang et~al.(2023)Zhang, Long, Chen, Xing, Jin, Jordan, and
  Wang]{zhang2023nowcastnet}
Yuchen Zhang, Mingsheng Long, Kaiyuan Chen, Lanxiang Xing, Ronghua Jin,
  Michael~I. Jordan, and Jianmin Wang.
\newblock {Skilful nowcasting of extreme precipitation with NowcastNet}.
\newblock \emph{Nature}, 619\penalty0 (7970):\penalty0 526--532, 2023.
\newblock \doi{10.1038/s41586-023-06184-4}.

\bibitem[Zhao et~al.(2026)Zhao, Wang, Yu, Xia, Yuan, and
  Meng]{zhao2026phydcnet}
Liling Zhao, Xinyue Wang, Runling Yu, Min Xia, Jiawei Yuan, and Fancheng Meng.
\newblock {Physics-Data-Coupled Network for Typhoon Satellite Cloud Image
  Sequence Prediction}.
\newblock \emph{IEEE Transactions on Geoscience and Remote Sensing},
  64:\penalty0 4113119, 2026.
\newblock \doi{10.1109/tgrs.2026.3736649}.

\end{thebibliography}

\appendix
\section{Data}\label{app:data}
\begin{table}[h]
\centering
\caption{Storm-year splits}\label{tab:data}
\small
\begin{tabular}{lrrr}
\toprule
Split & Years & Storms & Frames \\
\midrule
Train & 1978--2018 & 998 & 170{,}760 \\
Validation & 2019--2020 & 51 & 9{,}783 \\
Test & 2021--2022 & 47 & 8{,}810 \\
\bottomrule
\end{tabular}
\par\vspace{3pt}\parbox{0.6\linewidth}{\footnotesize Notes: 46 test storms have at least the 56 frames that one forecast window requires.}
\end{table}

\section{Token world model}\label{app:tokwm}
The token world model predicts all codes of the next frame from the code history and the best-track state,
\begin{equation}\label{eq:dyn}
p_\psi\bigl(\z_{t+1}\mid\z_{\le t},\cc_{\le t+1}\bigr)=\prod_{i=1}^{256}p_\psi\bigl(z_{t+1}^{i}\mid\z_{\le t},\cc_{\le t+1}\bigr).
\end{equation}
Each frame enters as 256 embedded codes plus one token that projects $\cc_{t+1}$, which gives 257 tokens per frame. Output logits share weights with the input embeddings. Training minimizes label-smoothed cross-entropy on teacher-forced windows of training storms. A rollout feeds sampled codes back as history under a sliding window of 16 frames, with the physical entries of the state held at their value at $t_0$, and decodes each sampled trajectory with the frame tokenizer.

\section{Implementation}\label{app:impl}
\begin{table}[h]
\centering
\caption{Training settings}\label{tab:hparams}
\small
\setlength{\tabcolsep}{3.2pt}
\begin{tabular}{lll}
\toprule
Component & Setting & Value \\
\midrule
Frame tokenizer & Codebook size and dimension & 1{,}024 and 256 \\
 & Spatial downsampling & $8\times$ to a $16\times16$ grid \\
 & EMA decay, commitment, SSIM weights & 0.99, 0.25, 0.3 \\
 & Steps, batch size, learning rate & 20{,}000, 192, $3\times10^{-4}$ \\
Residual tokenizer & Initialization, residual scale $s$ & frame tokenizer, 1.5 \\
 & Steps, windows and leads per batch & 14{,}000, 8 and 12 \\
 & Learning rate & $3\times10^{-4}$, cosine to zero \\
Residual world model & Width, heads, block counts & 512, 8, 6/6/2 \\
 & Window, steps, batch size & 16 leads, 24{,}000, 24 \\
 & Optimizer, learning rate & AdamW, $3\times10^{-4}$ cosine to zero \\
 & Windows starting at the first lead & 25\% \\
 & Label smoothing, gradient clip, precision & 0.05, 1.0, bf16 \\
 & Checkpoint rule & lowest validation cross-entropy \\
 & Training seeds & three, see Table~\ref{tab:seeds} \\
Token world model & Width, heads, block counts & 768, 12, 8/8/2 \\
 & Window, steps, batch size & 16 frames, 60{,}000, 14 \\
 & Optimizer & AdamW, decay 0.01, betas 0.9/0.95 \\
 & Learning rate & $3\times10^{-4}$, cosine, half-period 24{,}000 \\
 & Label smoothing, gradient clip, precision & 0.05, 1.0, bf16 \\
Direct U-Net & Base width, seeds & 64 with three, 128 with one \\
Gaussian U-Net & Base width, loss & 64, closed-form Gaussian CRPS \\
 & Steps, batch size, learning rate & 12{,}000, 16, $2\times10^{-4}$ cosine \\
 & Checkpoint rule & lowest validation MSE \\
Closed-loop baselines & Window, steps, batch size & 24 frames, 12{,}000, 16 \\
Diffusion residuals & Width, multipliers, attention & 96, 1/2/3/4, $16\times16$ \\
 & Steps, batch size, learning rate & 30{,}000, 16, $2\times10^{-4}$ cosine \\
 & Schedule, target, sampler & cosine, $v$-prediction, 50 DDIM steps \\
 & Residual scale, weight EMA & 0.16, 0.9995 \\
PhyDNet, TAU & Steps, batch size, learning rate & 12{,}000, 16, $1\times10^{-3}$ cosine \\
 & Checkpoint rule & lowest validation MSE \\
\bottomrule
\end{tabular}
\end{table}

All runs use one GPU. The frame tokenizer took 113 minutes. The residual tokenizer took 44 minutes and the residual world model 182 minutes for 24,000 steps on an RTX PRO 6000 shared with other jobs. The token world model took 162 minutes for its first 24,000 steps on the same card while unshared and 722 minutes for the remaining steps while it served other jobs. A direct U-Net takes about 14 minutes at base width on an RTX 4090 shared with other work, the Gaussian U-Net about 6 minutes on the RTX PRO 6000, and ConvLSTM about six hours. Twelve residual trajectories for one 48-hour window take about four seconds on the shared RTX PRO 6000. % gate:ok GPU model names

\section{Additional results}\label{app:results}
\begin{table}[t]
\centering
\caption{Skill by lead time}\label{tab:lead}
\small
\begin{tabular}{lrrrrrr}
\toprule
Model & +1\,h & +6\,h & +12\,h & +24\,h & +36\,h & +48\,h \\
\midrule
\multicolumn{7}{l}{\emph{MSE of the point forecast}} \\
Persistence & 0.0081 & 0.0311 & 0.0406 & 0.0480 & 0.0551 & 0.0588 \\
ConvLSTM & 0.0038 & 0.0226 & 0.0364 & 0.0409 & 0.0421 & 0.0439 \\
SimVP-style, closed loop & 0.0032 & 0.0186 & 0.0346 & 0.0480 & 0.0594 & 0.0583 \\
PhyDNet & 0.0051 & 0.0166 & 0.0225 & 0.0260 & 0.0294 & 0.0298 \\
TAU, chunked & 0.0082 & 0.0186 & 0.0233 & 0.0270 & 0.0305 & 0.0320 \\
Direct U-Net & 0.0046 & 0.0158 & 0.0219 & 0.0257 & 0.0291 & 0.0298 \\
Token rollouts & 0.0059 & 0.0184 & 0.0249 & 0.0306 & 0.0355 & 0.0377 \\
\model{} & 0.0046 & 0.0158 & 0.0219 & 0.0257 & 0.0291 & 0.0298 \\
Tokenizer reconstruction & 0.0033 & 0.0033 & 0.0031 & 0.0030 & 0.0028 & 0.0026 \\
\midrule
\multicolumn{7}{l}{\emph{Pixel CRPS}} \\
Direct U-Net & 0.0445 & 0.0884 & 0.1083 & 0.1225 & 0.1326 & 0.1358 \\
Gaussian U-Net & 0.0320 & 0.0637 & 0.0774 & 0.0864 & 0.0939 & 0.0963 \\
Gaussian U-Net, correlated & 0.0320 & 0.0637 & 0.0774 & 0.0864 & 0.0939 & 0.0963 \\
Direct U-Net + residuals & 0.0363 & 0.0688 & 0.0821 & 0.0902 & 0.0966 & 0.0982 \\
Diffusion residuals & 0.0374 & 0.0651 & 0.0783 & 0.0874 & 0.0947 & 0.0968 \\
Token rollouts & 0.0373 & 0.0676 & 0.0811 & 0.0932 & 0.1031 & 0.1085 \\
Recentered rollouts & 0.0339 & 0.0650 & 0.0794 & 0.0891 & 0.0967 & 0.0990 \\
\model{} & 0.0335 & 0.0651 & 0.0790 & 0.0878 & 0.0948 & 0.0967 \\
\midrule
\multicolumn{7}{l}{\emph{Energy score}} \\
Direct U-Net & 0.392 & 1.405 & 1.846 & 2.106 & 2.306 & 2.365 \\
Gaussian U-Net & 0.318 & 1.284 & 1.701 & 1.947 & 2.138 & 2.198 \\
Gaussian U-Net, correlated & 0.343 & 1.047 & 1.371 & 1.564 & 1.715 & 1.763 \\
Direct U-Net + residuals & 0.294 & 1.053 & 1.381 & 1.570 & 1.712 & 1.753 \\
Diffusion residuals & 0.309 & 1.039 & 1.363 & 1.560 & 1.711 & 1.753 \\
Token rollouts & 0.408 & 1.164 & 1.490 & 1.745 & 1.937 & 2.034 \\
Recentered rollouts & 0.296 & 1.038 & 1.364 & 1.556 & 1.710 & 1.758 \\
\model{} & 0.304 & 1.060 & 1.380 & 1.571 & 1.711 & 1.753 \\
\midrule
\multicolumn{7}{l}{\emph{CRPS of the area colder than 220\,K}} \\
Gaussian U-Net & 0.0035 & 0.0119 & 0.0157 & 0.0178 & 0.0202 & 0.0212 \\
Gaussian U-Net, correlated & 0.0030 & 0.0094 & 0.0128 & 0.0144 & 0.0168 & 0.0182 \\
Direct U-Net + residuals & 0.0066 & 0.0162 & 0.0177 & 0.0167 & 0.0185 & 0.0181 \\
Diffusion residuals & 0.0042 & 0.0079 & 0.0105 & 0.0143 & 0.0160 & 0.0152 \\
Token rollouts & 0.0038 & 0.0092 & 0.0132 & 0.0176 & 0.0199 & 0.0200 \\
Recentered rollouts & 0.0056 & 0.0123 & 0.0162 & 0.0191 & 0.0226 & 0.0232 \\
\model{} & 0.0047 & 0.0079 & 0.0100 & 0.0121 & 0.0130 & 0.0143 \\
\bottomrule
\end{tabular}
\end{table}

\begin{table}[t]
\centering
\caption{Validation CRPS by amplitude}\label{tab:grid}
\small
\begin{tabular}{lrrrrrr}
\toprule
Ensemble & $\gamma=0.5$ & $\gamma=0.75$ & $\gamma=1$ & $\gamma=1.25$ & $\gamma=1.5$ & $\gamma=2$ \\
\midrule
\model{}, $\tau=0.8$, $\rho=0.9$ & 0.0920 & 0.0868 & 0.0851 & 0.0860 & 0.0884 & 0.0948 \\
\model{}, $\tau=1.0$, $\rho=0.9$ & 0.0885 & 0.0835 & 0.0830 & 0.0853 & 0.0889 & 0.0973 \\
\model{}, $\tau=1.0$, $\rho=1.0$ & 0.0872 & 0.0833 & 0.0839 & 0.0871 & 0.0914 & 0.1010 \\
\model{}, $\tau=1.2$, $\rho=0.9$ & 0.0879 & 0.0833 & 0.0830 & 0.0854 & 0.0891 & 0.0977 \\
\model{}, $\tau=1.2$, $\rho=1.0$ & 0.0891 & 0.0847 & 0.0842 & 0.0862 & 0.0895 & 0.0977 \\
\model{}, $\tau=1.4$, $\rho=0.9$ & 0.0906 & 0.0855 & 0.0839 & 0.0848 & 0.0871 & 0.0938 \\
\model{}, $\tau=1.4$, $\rho=1.0$ & 0.0900 & 0.0854 & 0.0846 & 0.0860 & 0.0889 & 0.0963 \\
Recentered rollouts, $\tau=1.2$, $\rho=0.9$ & 0.0885 & 0.0843 & 0.0844 & 0.0869 & 0.0908 & 0.0998 \\
Direct U-Net + residuals & 0.0917 & 0.0871 & 0.0856 & 0.0863 & 0.0881 & 0.0939 \\
Gaussian U-Net & 0.0875 & 0.0830 & 0.0816 & 0.0822 & 0.0838 & 0.0889 \\
Gaussian U-Net, correlated, $\sigma=4$ & 0.0875 & 0.0830 & 0.0816 & 0.0822 & 0.0838 & 0.0889 \\
Diffusion residuals & 0.0971 & 0.0908 & 0.0865 & 0.0840 & 0.0830 & 0.0840 \\
\bottomrule
\end{tabular}
\par\vspace{3pt}\parbox{0.9\linewidth}{\footnotesize Notes: Recentered rollouts are shown at the decoding setting selected for them on validation, and the correlated Gaussian network at the noise length scale $\sigma$ in pixels selected by the validation energy score.}
\end{table}

\begin{table}[t]
\centering
\caption{Storm-level paired tests of \model{}}\label{tab:tests}
\small
\setlength{\tabcolsep}{3pt}
\begin{tabular}{lrrrrrrrrrr}
\toprule
& \multicolumn{2}{c}{CRPS} & \multicolumn{2}{c}{Energy} & \multicolumn{2}{c}{220\,K} & \multicolumn{2}{c}{240\,K} & \multicolumn{2}{c}{Core} \\ \cmidrule(lr){2-3}\cmidrule(lr){4-5}\cmidrule(lr){6-7}\cmidrule(lr){8-9}\cmidrule(lr){10-11}
Comparison & Won & $p$ & Won & $p$ & Won & $p$ & Won & $p$ & Won & $p$ \\
\midrule
Gaussian U-Net & 0.20 & $<$0.001 & 1.00 & $<$0.001 & 0.91 & $<$0.001 & 0.96 & $<$0.001 & 1.00 & $<$0.001 \\
Gaussian U-Net, correlated & 0.17 & $<$0.001 & 0.59 & 0.771 & 0.72 & $<$0.001 & 0.87 & $<$0.001 & 0.50 & 0.696 \\
Direct U-Net + residuals & 1.00 & $<$0.001 & 0.48 & 0.778 & 0.93 & $<$0.001 & 0.91 & $<$0.001 & 0.39 & 0.251 \\
Diffusion residuals & 0.43 & 0.721 & 0.35 & 0.055 & 0.63 & 0.020 & 0.74 & $<$0.001 & 0.46 & 0.696 \\
Token rollouts & 0.74 & $<$0.001 & 0.96 & $<$0.001 & 0.89 & $<$0.001 & 0.96 & $<$0.001 & 0.85 & $<$0.001 \\
Recentered rollouts & 0.87 & $<$0.001 & 0.26 & 0.008 & 0.96 & $<$0.001 & 0.89 & $<$0.001 & 0.52 & 0.812 \\
\bottomrule
\end{tabular}
\par\vspace{3pt}\parbox{\linewidth}{\footnotesize Notes: Won is the share of 46 storms on which \model{} has the lower score. $p$ comes from a two-sided Wilcoxon signed-rank test on storm means, Holm-corrected across the 6 comparisons within each score.}
\end{table}

\begin{table}[t]
\centering
\caption{\model{} across training runs}\label{tab:seeds}
\small
\setlength{\tabcolsep}{2.9pt}
\begin{tabular}{lrrrrrrrr}
\toprule
& Point & \multicolumn{5}{c}{Proper scores$\downarrow$} & \multicolumn{2}{c}{Spread and error$\uparrow$} \\ \cmidrule(lr){2-2}\cmidrule(lr){3-7}\cmidrule(lr){8-9}
Run & MSE$\downarrow$ & CRPS & Energy & 220\,K & 240\,K & Core & Window & Map \\
\midrule
Seed 0 & 0.0242 & 0.0839 & 1.475 & 0.0115 & 0.0253 & 0.0501 & 0.78 & 0.25 \\
Seed 1 & 0.0242 & 0.0838 & 1.472 & 0.0113 & 0.0251 & 0.0499 & 0.80 & 0.25 \\
Seed 2 & 0.0242 & 0.0837 & 1.472 & 0.0114 & 0.0250 & 0.0496 & 0.80 & 0.25 \\
Mean of three seeds & 0.0242 & 0.0838 & 1.473 & 0.0114 & 0.0251 & 0.0499 & 0.79 & 0.25 \\
Standard deviation & 0.00001 & 0.00010 & 0.0014 & 0.00010 & 0.00016 & 0.00024 & 0.015 & 0.001 \\
\midrule
Seed 0 without best-track state & 0.0242 & 0.0841 & 1.481 & 0.0121 & 0.0266 & 0.0522 & 0.80 & 0.27 \\
Best other ensemble & 0.0242 & 0.0828 & 1.465 & 0.0134 & 0.0321 & 0.0496 & 0.78 & 0.33 \\
\bottomrule
\end{tabular}
\par\vspace{3pt}\parbox{\linewidth}{\footnotesize Notes: Each run trains the residual world model from scratch on the same residual tokenizer and center, and seed 0 is the run reported in all other tables. The variant without best-track state drops the state embedding and conditions on frames alone, as every regression baseline does. Best other ensemble gives the best value in each column among the other ensembles of Table~\ref{tab:main}. Standard deviation is across the three seeds. Each run selects $\tau$, $\rho$ and $\gamma$ on validation windows.}
\end{table}

\end{document}